\documentclass[10pt,conference,a4paper]{IEEEtran}
\usepackage{cite}

\ifCLASSINFOpdf
  \usepackage[pdftex]{graphicx}
  \graphicspath{{./img/}}
\else
\fi

\usepackage{amsmath}

\usepackage{array}

\ifCLASSOPTIONcompsoc
  \usepackage[caption=false,font=normalsize,labelfont=sf,textfont=sf]{subfig}
\else
  \usepackage[caption=false,font=footnotesize]{subfig}
\fi

\usepackage{hyperref}

\usepackage{eurosym}
\usepackage{multirow}
\usepackage[table]{xcolor}
\usepackage{amsthm}

\begin{document}
%
\title{A Novel Approach for Data-Driven Automatic Site Recommendation and Selection}



%
\author{\IEEEauthorblockN{
		Sebastian Baumbach$^{1,2}$,
		Frank Wittich$^2$,
		Florian Sachs$^3$, 
		Sheraz Ahmed$^1$ and
		Andreas Dengel$^{1,2}$}
	\IEEEauthorblockA{$^1$German Research Center for Artificial 		
		Intelligence, Kaiserslautern, Germany}
	\IEEEauthorblockA{$^2$University of Kaiserslautern, Germany}
	\IEEEauthorblockA{$^3$University of Magdeburg,
		Faculty of Economics and Management, Magdeburg, Germany}
	\IEEEauthorblockA{Email: sebastian.baumbach@dfki.de}
}


\maketitle
\begin{abstract}
This paper presents a novel, generic, and automatic method for data-driven site selection. Site selection is one of the most crucial and important decisions made by any company. 
Such a decision depends on various factors of sites, including socio-economic, geographical, ecological, as well as specific requirements of companies.
The existing approaches for site selection (commonly used by economists) are manual, subjective, and not scalable, especially to Big Data. The presented method for site selection is robust, efficient, scalable, and is capable of handling challenges emerging in Big Data. 
To assess the effectiveness of the presented method, it is evaluated on real data (collected from Federal Statistical Office of Germany) of around 200 influencing factors which are considered by economists for site selection of Supermarkets in Germany (Lidl, EDEKA, and NP). Evaluation results show that there is a big overlap (86.4 \%) between the sites of existing supermarkets and the sites recommended by the presented method. In addition, the method also recommends many sites (328) for supermarket where a store should be opened.
\end{abstract}


%
\IEEEpeerreviewmaketitle

\section{Introduction}
\label{sec:introduction}

Selecting a facility location is a constitutive investment decision of critical importance for every company that wants to operate a successful business \cite{Strotmann2007, Bhatnagar2005443}. Every company faces it at least once in its lifetime. Negative influences of a sites' location can hardly be compromised by other factors or actions \cite{woratschek2000, glatte2015location}. Numerous geographical, social, economic, or socio-economic factors for each site are usually be taken into account and compared against the specific requirements of a company \cite{blair1987major}.
\newline
In the modern era of rapid data collection and Big Data, countless information sources about locations are easily available. They cover not only detailed maps like Google Maps or OpenStreetMap, but also data about demographical and social attributes of inhabitants, which is, for instance, offered by the Federal Statistical Office of Germany as open data. In addition to open data, there exist many companies focusing on the creation of complex datasets, ranging from regional purchasing power distribution to consumer behavior of citizens. As a consequence, rich and heterogeneous datasets are available, containing spatial and temporal information that is of high relevance to decide where to locate a site.
\newline
In the area of site selection, research work traces back nearly a century \cite{lehr1885mathematische, mcmillan1965manufacturers, weber1909urber}. However, due to globalization and digitization, geographical search space as well as amount of data has enormously increased. This exponential growth of decisive data makes it increasingly difficult for decision-makers and experts to analyze all relevant information manually. This is one way why locations are selected from \textit{top} (i.e. state) to \textit{bottom} (i.e. region) in practice. As a consequence, other regions are excluded and the selection is commonly based on subjective criteria \cite{mcmillan1965manufacturers, luder1983unternehmerische, Zelenovic2003}. The remaining sites are weighted with respect to their descriptive attributes and the "best" site wins this election. However, weighting and selecting are highly subjective and often based on the individual experience and assessment of the involved experts \cite{olbert1976standortentscheidungsprozess, schober1990strukturierte, greiner1997standortbewertung, rikalovic2015fuzzy}. 
\newline
The aim of this paper is to impartially find a suitable site based on analyzing decisive data of all influencing factors. Based on the assumption that site selection can be modeled quantitatively, all locations as well as the companies' requirements are treated respectively. Thus, sites are described by its attributes, e.g. \textit{purchasing power}, \textit{number of inhabitant} or \textit{proximity to suppliers}, which are called location factors (LF) in economics. A companies' requirements are expressed by a combination of these LF and appropriate weights. Based on this idea, a suitable site is a location where its corresponding LF fulfill the requirements of a specific company. This leads to the research questions a) how to build the required data model and b) how to extract these patterns from the datasets? The utilization of knowledge-based decision-making and Big Data for supporting companies in their site selection process is new in the area of location analysis.
\newline
This paper presents a novel approach for data-driven site selection. It combines a data model for aggregating LF with an enhanced recommendation system. The data model captures LF by incorporating spatio-temporal data from different sources and structuring it in a geographical, temporal, and hierarchical way. These sites and their data, representing items in the context of recommender systems, face user requirements that are made up of constraints, preferences and weights. They are matched by using a means-end relation between the corresponding data and companies' requirements. Thereby, its flexibility allows the system to be used in many different industries with diverse requirements.
\newline
The contribution of this paper is an approach for data-driven site selection, which analyzes location factors and company requirements (Section \ref{sec:methodology}). It is built upon a hierarchical data model and combines knowledge about site selection with extensions to recommender systems. A case study about supermarkets is also conducted in order to demonstrate the viability of the presented approach (Section \ref{sec:evaluation}). Therefore, the actual locations of all supermarkets in Germany were compared with the suggestions for suitable places made by this system. The recommender system was configured with the requirements for supermarket, which were taken from economic literature (\ref{sec:conclusion}). Finally, the results show that this approach is not only able to suggest sites to companies, but also helps to identify unknown pattern behind locations and its characteristics.

\section{State-of-the-Art in Locational Datasets, Economics, \& Recommender Systems}
\label{sec:start_of_the_art}

The existing system commonly used for recommending movies, books, or music \cite{lee2010collaborative, carrer2012social, nunez2012implicit} cannot be used for site selection. This is because, this method cannot be configured to incorporate application context with a complexity similar to location analysis. The existing methods used by economists \cite{Zelenovic2003, woratschek2004dienstleistungsmanagement, arauzo2010empirical} are very complex and time extensive (as they require a lot of manual configuration and analysis), which limit their applicability in practice.

Investigations of sites have been only made on base of small datasets, e.g. population density, skill level of human capital, and infrastructure of 720 municipalities as studied by Arauzo Carod \cite{arauzo2005determinants}.


\subsection{Site Selection}
\label{sec:SiteSelection}

Analyzing locations is done by comparing the LF of different places. Location factors are the properties that are influential towards a companies goal achievement \cite{liebmann1969grundlagen}. The important LF are usually chosen by a company based on their own demands and each company might posses different requirement. Table \ref{tab:LocationFactors} shows the most commonly LF grouped into 10 major categories \cite{badri2007dimensions}. 

\begin{table}[!ht]
	\def\arraystretch{1.25}
	\caption{Categorized Location Factors, Selection}
	\label{tab:LocationFactors}
	\centering
	\resizebox{1\columnwidth}{!}{
	\begin{tabular}{l|p{7.5cm}}
		\textbf{Category} & \textbf{Location Factor} \\ 
		\hline
		Transportation & Highways, railroad, waterways, airways. \\ 
		\hline
		Labor & Availability, educational level \& wage rates of labor. \\ 
		\hline
		Raw Materials & Proximity to supplies, closeness to component parts. \\ 
		\hline
		Markets & Existing market, growth of markets, competitors, trends. \\ 
		\hline
		Industrial Site & Accessibility of land, space for future expansion. \\ 
		 \hline
		Utilities & Disposable facilities, availability of fuels, electric power \& gas. \\ 
		\hline
		Tax Structure & Industrial property tax rates, tax free operations, state sales tax. \\ 
		\hline
		Climate & Amount snow fall \& rain fall, average temperature, air pollution. \\ 
		\hline
		Community & Colleges and research institutions, quality of schools. \\ 
	\end{tabular}
	}
\end{table}

Based on these LF, there are many ways how site selection can be done as there is no fixed procedure. However, there are few guidelines and operating procedures how a site should be selected.

In literature and practice, the site selection process is often divided into multiple phases where the regional focus is reduced in each phase. Zelenovic split it up into a macro and micro selection \cite{Zelenovic2003}. The first phase addresses the issue of finding the right state while the second one looks for a specific site within the previously chosen state. Bankhofer divided the whole process into four phases of selecting continent, country, municipality and then final location in that order \cite{Bankhofer2001}. However, this division "top-down" is impartial and inefficient as it requires manual analysis and selection. The exploration and selection of all decisive LF by hand is not feasible since the amount of data is increasing exponentially.

Models like Discrete Choice Models or Count Data Model for weighting and selection have become increasingly complex over the years \cite{arauzo2010empirical}. They include not only numerous, but also constantly changing LF based on the varying contexts of each model. There is no similar study which reveals identical findings, but suggests different subsets of LF \cite{blair1987major}. Even though these models are theoretically helpful, they are too complex and therefore, time and cost intensive in a real scenario. This leads managers to base their decisions not only on given facts, but rather on their personal and emotional judgment \cite{olbert1976standortentscheidungsprozess, schober1990strukturierte, greiner1997standortbewertung, rikalovic2015fuzzy}.

Woratschek and Pastowski studied different methods in economic used for location selection \cite{woratschek2004dienstleistungsmanagement}. Checklist Methods are a very basic way to assist the selection process. Relevant LF for a company get listed and weighted by experts for different locations. These weights are the degree how good a location fulfill a given requirement. Extensions are Selection by Elimination, which immediately removes location which do not fulfill a requirement, and Scoring Models, which rates the weighted LF by a given user-defined scale. The result of these methods are only qualitative nature and do not depend on quantitative findings. Furthermore, these ranked lists do not provide any in-depths analysis. 

\subsection{Recommender Systems}
\label{sec:RecommenderSystem}

Traditionally, items are defined by their attributes, which contain keywords or numbers. Users are described by their preferences which basically express their interest in certain items. A detailed introduction to recommender systems can be found in \cite{nextGen}. However, this basic model for recommendation is not suitable for site selection, because:
\begin{itemize}
	\item For suggesting suitable sites to companies, different forms of knowledge have to be utilized which exceeds the basic modeling of users and items. This data covers preferences, constraints, context, and domain knowledge about the users' needs, others users, the locations themselves and past recommendations \cite{felfernig2008constraint}.
	\item Most of the current recommender systems for books \cite{nunez2012implicit}, movies \cite{carrer2012social} or music \cite{lee2010collaborative} deals with single-criterion ratings, such as ratings of movies. However, in site selection, it is essential to integrate \textit{multi-criteria ratings}, e.g., a company can have different constraints, which must be fulfilled, as well as multiple preferences, which should be satisfied \cite{nextGen}.
\end{itemize}

\section{The proposed Approach for Site Selection}
\label{sec:methodology}

\begin{figure}[!ht]
	\centering
	\includegraphics[width=\columnwidth]{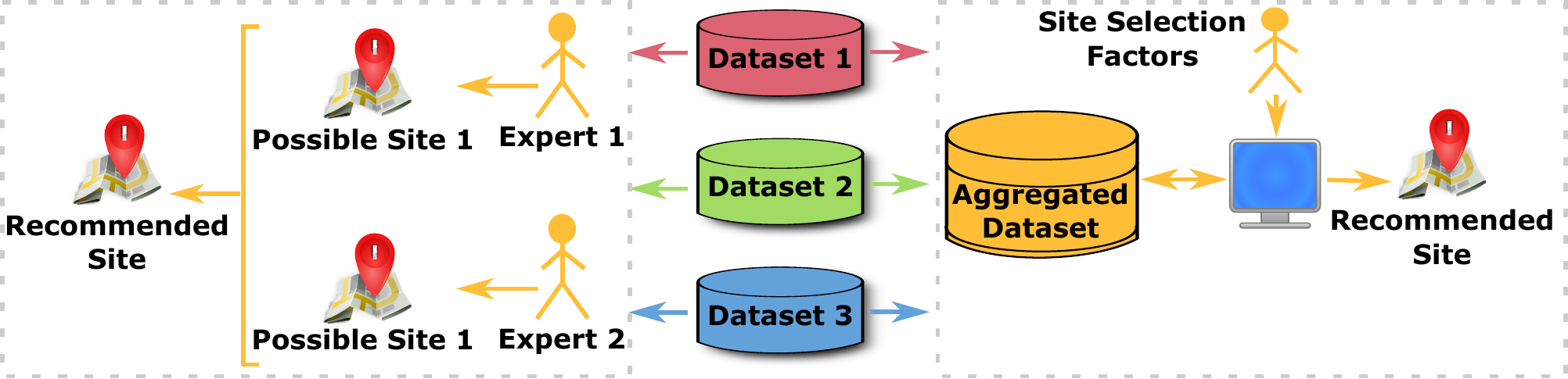}
	\caption{Site Selection: Existing vs. proposed Site Selection.}
	\label{fig:Comparison_Approaches}
\end{figure}

The proposed approach compares companies' requirements with locations based on their LF and simultaneously, utilizes the available knowledge concerning locations and companies' requirements. It turns the \emph{manual exploration of possible sites by decision-makers and experts} into a \emph{definition of requirements by decision-makers \& getting recommendations by the expert system}.  Figure \ref{fig:Comparison_Approaches} shows the comparison between traditional and data-driven site selection..

\begin{figure}[htbp]
	\centering
		\includegraphics[width=\columnwidth]{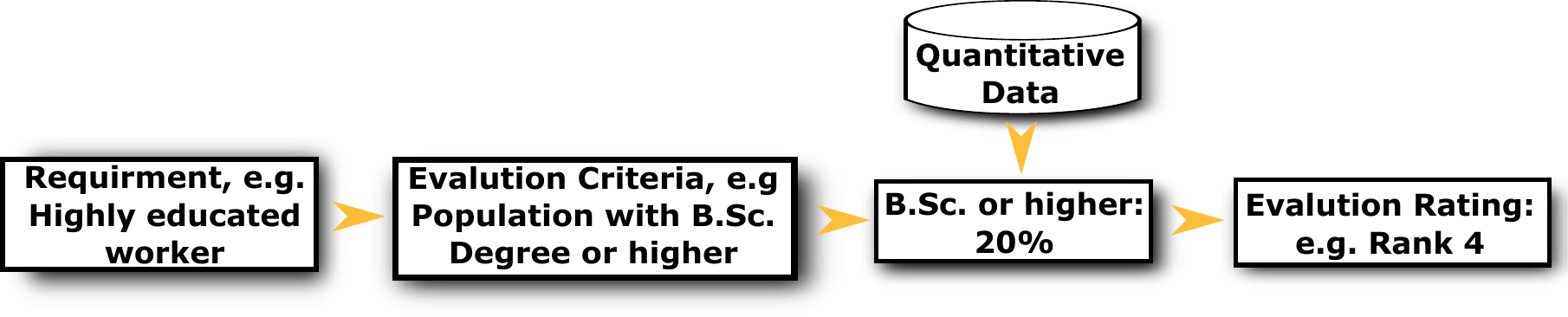}
	\caption{Requirements, Quantitative Data, and Qualitative Evaluation.}
	\label{fig:RequirementProcess}
\end{figure}

The most important problem in site selection is that decisions should be made based on qualitative measurement and evaluation. However, data available is mostly quantitative. Inspired by the recommendation methods used by economists, in the presented automatic recommendation method all requirements are first prioritized and scored based on their relevance and importance for the company. Figure \ref{fig:RequirementProcess} illustrates the requirements are then used to rank different locations.

\begin{figure}[htbp]
	\centering
	\includegraphics[width=\columnwidth]{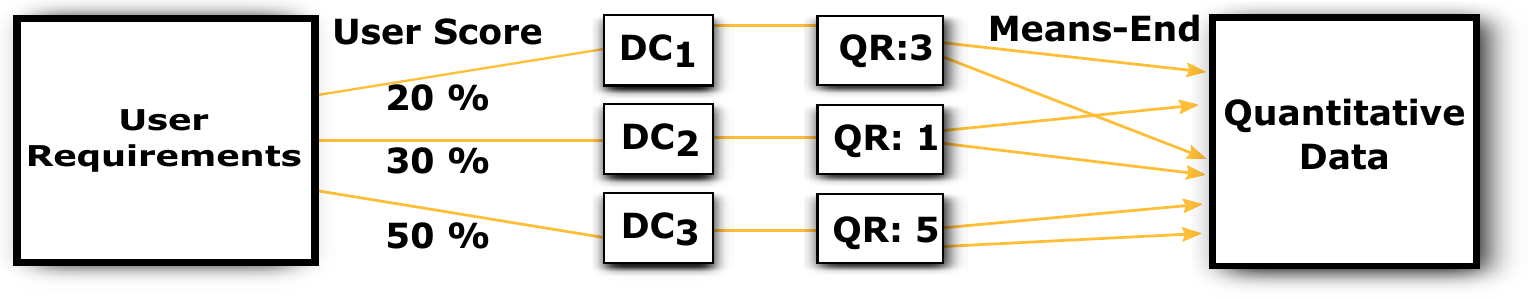}
	\caption{Abstraction of the URP Model.}
	\label{fig:URP_Abstraction}
\end{figure}

The User Requirements Profile (URP) is a composition of the Decision Criteria (DC) and Qualitative Ratings (QR). DC represents a company's requirements for a new location. QR is the evaluation of a criteria based on the quantitative data and thus, used to make qualitative suggestions. In this sense, a content-based-filtering is utilized. Figure \ref{fig:URP_Abstraction} displays the URP model. It incorporates ideas and concepts from Felfernig and Burke in which, among others, constraints, preferences and means-ends are applied in a recommender system \cite{felfernig2008constraint}.

DC is the top-level components of URP, representing abstract requirements like \textit{good infrastructure} or \textit{weak competitors}. DC also utilize the concept of scoring and weighting of LF originally introduced by scoring models (see Sec. \ref{sec:SiteSelection}). Additionally, the concept of 'Selection by Elimination"' were adopted where locations get eliminated if they do not fulfill all of the requirements. Each DC can either be classified as 'Must-Have' and thus, representing an elimination condition if this criteria is not fulfilled, or as constraint in form of 'Preferences' or respectively 'Nice-To-Haves'.
\newline
Each QR delivers an evaluation for the corresponding DC based on the LF stored in the data model. Companies' needs cannot always be directly translated to LF. This is the reason why fuzzification has been applied \cite{cao2007MeansEnd}. A LF might satisfy different quality needs and in return a qualitative need encompasses multiple attributes. The proposed approach utilizes means-end relations as introduced in the work Felfernig and Burke \cite{felfernig2008constraint}, which allows reasoning about "how particular items (the means) satisfy particular needs or requirements of the user (the ends)". Ultimately, they defined the recommendation task as a \textit{constraint satisfaction problem}, which has been adopted to site selection.  
Given a set of the company's requirements, the recommendation can be calculated by using a constraint-based recommender $R_{constr}$ which computes solutions \{$site_i \in AllSites$\} for a given site recommendation task.

\newtheorem{mydef}{Definition}

\begin{mydef}
	\label{def:CSP}
	Based on \cite{felfernig2008constraint}, a site recommendation task is defined as Constraint Satisfaction Problem $(C, S, CR \cup COMP \cup FILT \cup SITES)$ where $C$ is a finite set of variables representing potential requirements of the company and $S$ is a set of variables defining the  basic properties of the sites. Furthermore, CR is a set of company's requirements,	$COMP$ represents a set of (incompatibility) constraints, $FILT$ is a set of filter constraints, and $PROD$	specifies the set of offered sites. 
\end{mydef}

A solution to a given site recommendation task $(C, S, CR \cup COMP \cup FILT \cup SITES)$ is a complete assignment to the variables of $(C, S)$ such that this assignment  is consistent with the constraints in $(CR \cup COMP \cup FILT \cup SITES)$. A weighting can be defined on those relations to fine tune the relevance of every LF according the companies' requirements.

\section{Evaluation}
\label{sec:evaluation}

For the evaluation, more than 200 datasets have been collected and imported into the data model. The conducted experiments focus on supermarkets in Germany as one possible application context for which the relevant LF has been provided by experts \cite{greiner1997standortbewertung}. However, the presented approach is generic and can be applied to datasets collected from any country as well as to a wide range of problems, e.g., gas stations, cinemas, gold or oil mines.

\subsection{Hierarchical Data Model}

In this paper, the recommendation system in evaluated on data collected in Germany. 
The administration hierarchy contains Germany with its 16 states, 402 countries, 4,520 districts, and 11,162 municipalities as well as information about the hierarchy. As for now, municipalities are the lowest level of territorial division. This information is provided by the \href{https://www.destatis.de/DE/ZahlenFakten/LaenderRegionen/Regionales/Gemeindeverzeichnis/Administrativ/AdministrativeUebersicht.html;jsessionid=89930EA06D623C492E12165EBAD494FD.cae4}{Federal Statistical Office of Germany} and can be used without any restrictions and free of charge. 


Data for over 200 LF have been integrated into the data model, which is also provided by the \href{https://www.regionalstatistik.de}{Federal Statistical Office}. Available LF cover topics like net income, purchasing power, various information about inhabitants \& population, employees \& unemployed persons, education, land costs, households in number \& size, and companies itemized in size, number, profit, industry, \& employed people. A selection of most decisive LF\footnote{The LF have been chosen based on the results of Section \ref{subsec:attribute_selection}.} for Berlin is provided in Table \ref{tab:berlin}. Ultimately, there are values for each of the 200 LF, over the last 15 till 20 years (depending on the specific LF), and all sites (the sum of all states, counties, districts, and municipalities) in total. In this paper, the latest values for the current year 2016 have been used.

\begin{table}[!ht]
	\def\arraystretch{1.25}
	\caption{Selected Location Factors for Berlin.}
	\label{tab:berlin}
	\centering
	\resizebox{1\columnwidth}{!}{
		\begin{tabular}{p{6cm}|r}
			\textbf{Factor} & \textbf{Value} \\ 
			\hline
			Avg. GDP per Employee & \EUR{71,209} \\ 
			\hline
			Available income per Inhabitant & \EUR{22,586} \\ 
			\hline
			Inhabitant & 3,484,995 \\ 
			\hline
			Employment Rate & 78.9 \% \\
		\end{tabular}
	}
\end{table}

\begin{figure}[!ht]
	\centering
	\includegraphics[width=\columnwidth]{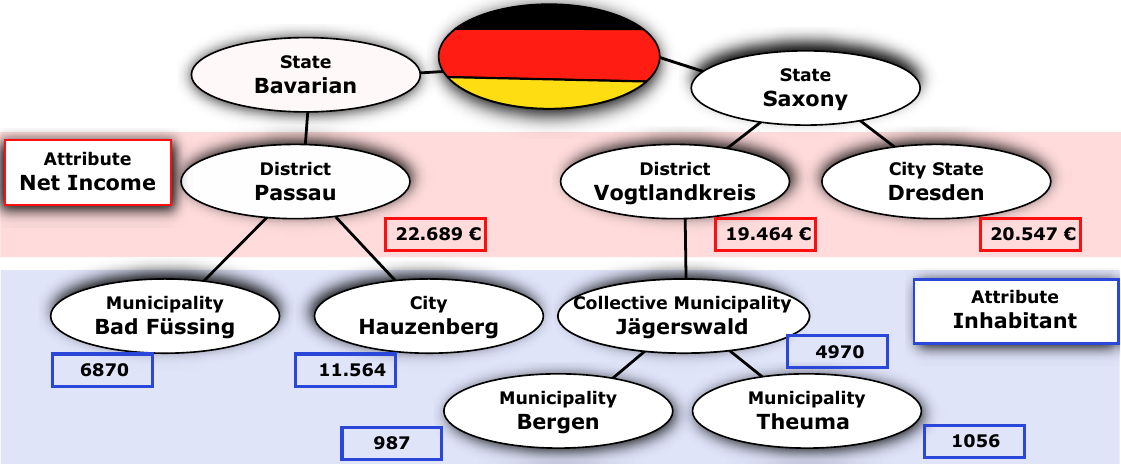}
	\caption{Illustration of Hierarchical Data Model.}
	\label{fig:Municipalities}
\end{figure}

Most of the LF are available for districts and some have been created on the granularity level of municipalities (see \ref{fig:Municipalities}). Each LF exit (e.g. net income) for all sites within the specific hierarchy level it has been created for (e.g. districts). LF are inherent up- and downwards: the inhabitants of a district (blue) is the sum of inhabitants of all municipalities whereas the average net income (red) for a district is also assigned to all of its municipalities and cities. In case of the LF \textit{net income}, there exit 402 values for all countries in the data model. 

\subsection{Application Focus}
\label{subsec:application}

Greiner, in charge of location selection in the \emph{REWE} Group analyzed in depth how this decision is performed and which LF are of interest for \emph{REWE} supermarkets \cite{greiner1997standortbewertung}. \emph{REWE} mainly uses the LF \textit{number of inhabitants} in the macro selection phase for selection municipalities in Germany. Beside this factor, the availability of parking places as well as real estate of certain size is another requirement which is, however, beyond the scope of this paper\footnote{The focus of this work lies on macro selection, whereas facility analysis is part of the micro selection (see Sec. \ref{sec:start_of_the_art}).}. In the second step, experts directly visit and assess the suitability of the pre-selected places. According to their expansion strategies, \emph{Edeka}, \emph{Lidl}, and \emph{NP} supermarket chains assert to apply the same criteria\footnote{Source: \href{http://www.lidl-immobilien.de/cps/rde/xchg/SID-6DF6A32A-8E7E665B/lidl_ji/hs.xsl/5186.htm}{www.lidl.de} and\href{https://www.edeka-verbund.de/Unternehmen/de/edeka_minden_hannover/immobilien_minden_hannover/expansion_minden_hannover/edeka_expansion_minden_hannover.jsp?}{www.edeka-verbund.de}.}.

\subsection{Comparison of Recommendations with Actual Locations}
\label{subsec:comparison}

86.4 \% of all locations can be explained by only taking these LF into account (Tab. \ref{tab:Overlap}). The requirements for suitable locations for supermarkets as described in the economic literature have been used to configure the recommender system and its URP. This profile contains the \textit{number of inhabitants} and the \textit{regional focus}. \emph{NP} is looking for sites where more than 2,500 inhabitants are living, whereas \emph{Edeka} and \emph{Lidl} are searching for locations where at least 5,000 inhabitants are living. \emph{E-Center}, which are stores times larger than the ones mentioned beforehand, are seeking for places with more than 10,000 inhabitants. Finally, the regional focus has been set to the states Saxony-Anhalt, Brandenburg, and Lower Saxony as the chain \emph{NP} operates in these states only. Within this regional focus, there exist 1,704 municipalities respectively cities with currently 481 \emph{Edeka}, 90 \emph{E-Center}, 256 \emph{NP}, and 453 \emph{Lidl} stores (sum 1,280).  

\begin{table}[!ht]
	\def\arraystretch{1.25}
	\caption{Overlap between existing Supermarket \& Recommended Sites.}
	\label{tab:Overlap}
	\centering
	\resizebox{1\columnwidth}{!}{
		\begin{tabular}{cr|c|c|c|c|c}
			\multicolumn{2}{c|}{} & \textbf{Edeka} & \textbf{E-Center} & \textbf{Lidl} & \textbf{NP} & Overall \\
			\hline
			\multicolumn{2}{c|}{No. of existing Supermarket} & 481 & 90 & 453 & 256 & 1280 \\  
			\hline
			\multirow{2}{*}{\textbf{Recommended Sites}} & No. & 364 & 80 & 428 & 224 & 1096 \\
			& Percentage & 74.8 \% & 88.8 \% & 94.5 \% & 87.5 \% &  86.4 \% \\
		\end{tabular}
	}
\end{table}

Additionally, 328 suitable sites were found where currently no supermarket of competitors is located and which can be recommended (51 to \emph{Lidl}, 88 to \emph{Edeka}, 140 to \emph{NP}, and 49 to \emph{E-Center}).

\begin{table}[!ht]
	\def\arraystretch{1.25}
	\caption{Recommended Sites for new Supermarket Stores.}
	\label{tab:Recommendations}
	\centering
	\resizebox{1\columnwidth}{!}{
		\begin{tabular}{cr|c|c|c|c}
			\multicolumn{2}{c|}{} & \textbf{Edeka} & \textbf{E-Center} & \textbf{Lidl} & \textbf{NP} \\
			\hline
			\multirow{2}{*}{\textbf{Recommended Sites}} & Total & 270 & 303 & 412 & 559\\
			& Without Markets & 88 & 49 & 51 & 140 \\
		\end{tabular}
	}
\end{table}

Table \ref{tab:Filialen} shows the distribution of supermarket stores. 94.5 \% of the \emph{Lidl}, 74.8 \%  of \emph{Edeka}, 87.5 \% of \emph{NP} and 88.8 \% of \emph{E-Center} stores are located at municipalities which fulfill the LF \textit{number of inhabitants}. Only in 25 (5.5 \%) cases, \emph{Lidl} opened a store at a site with less than 5,000 inhabitants. 23.3 \% and 11.1 \% of \emph{Edeka} respectively \emph{E-Center} stores can be found at sites which has less inhabitants than actually required. \emph{NP} has selected in 32 cases (12.5 \%) a municipality with less than 2,500 inhabitants.

\begin{table}[!ht]
	\def\arraystretch{1.25}
	\caption{Relation of the Supermarket Chains to the No. of Inhabitant.}
	\label{tab:Filialen}
	\centering
	\resizebox{1\columnwidth}{!}{
		\begin{tabular}{cc|p{1.5cm}|p{1.5cm}|c}
			\multicolumn{2}{c|}{\multirow{2}{*}{Store}} & \multicolumn{2}{c|}{\textbf{Criteria Inhabitants Fulfilled}} & \multirow{2}{*}{Overall} \\
			& & yes & no & \\
			\hline
			\multirow{2}{*}{\textbf{Lidl}} & yes & 428 & 25 & 453 \\   
			& no & 412 & 839 & 1251 \\
			\hline
			\multicolumn{2}{c|}{Overall} & 840 & 864 & 1704 \\
			\hline
			\hline
			\multirow{2}{*}{\textbf{Edeka}} & yes & 364 & 117 & 481 \\   
			& no & 270 & 953 & 1223 \\
			\hline
			\multicolumn{2}{c|}{Overall} & 634 & 1070 & 1704 \\
			\hline	
			\hline
			\multirow{2}{*}{\textbf{E-Center}} & yes & 80 & 10 & 90 \\   
			& no & 303 & 1311 & 1614 \\
			\hline
			\multicolumn{2}{c|}{Overall} & 383 & 1321 & 1704 \\	
			\hline	
			\hline
			\multirow{2}{*}{\textbf{NP}} & yes & 224 & 32 & 256 \\   
			& no & 599 & 849 & 1448 \\
			\hline
			\multicolumn{2}{c|}{Overall} & 783 & 881 & 1704 \\				
		\end{tabular}
	}
\end{table}


\subsection{Attribute Selection}
\label{subsec:attribute_selection}

Since these LF, as described by experts and supermarket companies themselves, are precise to a high degree, the follow-up question is: Are there any more decisive LF, irrespective of whether or not they are consciously used by the companies?
\newline
By means of a correlation analysis
, all remaining LF (more than 200) have been reviewed regarding their impact on the availability of supermarkets. The LF with the highest correlation scores are presented in Table \ref{tab:correlation}. It can be seen that, beside the dependency on the number of inhabitants, the purchasing power has an impact on the existence of supermarkets. However, this dependency is only high for \emph{Edeka} and \emph{Lidl}, which are, additionally, correlating among each other. Furthermore, it can be observed that \emph{NP} neither attach importance to the inhabitants nor the purchasing power in the way \emph{Edeka} and \emph{Lidl} set value on it.  

\begin{table}[!ht]
	\def\arraystretch{1.25}
	\caption{Correlation Matrix for Location Factors. $0$ (No Correlation) and $+1$ (High orrelation)}
	\label{tab:correlation}
	\centering
	\resizebox{1\columnwidth}{!}{
		\begin{tabular}{l|l|l|l|l}
			Attribute & Num. of Edeka & Num. of E-Center & Num. of Lidl & Num. of NP \\
			\hline
			Num. of Edeka Stores & \cellcolor{blue!50} 1.0 & \cellcolor{blue!30} 0.61 & \cellcolor{blue!47} 0.94 & \cellcolor{blue!24} 0.48 \\
			\hline 
			Purchasing Power & \cellcolor{blue!47} 0.95 & \cellcolor{blue!30} 0.61 & \cellcolor{blue!49} 0.99 & \cellcolor{blue!24} 0.47 \\
			\hline
			Num. of Inhabitants & \cellcolor{blue!47} 0.94 & \cellcolor{blue!30} 0.60 & \cellcolor{blue!49} 0.99 & \cellcolor{blue!23} 0.46 \\
			\hline
			Num. of Households & \cellcolor{blue!47} 0.94 & \cellcolor{blue!30} 0.60 & \cellcolor{blue!49} 0.99 & \cellcolor{blue!23} 0.46 \\
			\hline
			Num. of Lidl stores & \cellcolor{blue!47} 0.94 & \cellcolor{blue!30} 0.59 & \cellcolor{blue!50} 1.0 & \cellcolor{blue!22} 0.45 \\
			\hline
			Num. of E-Center & \cellcolor{blue!30} 0.61 & \cellcolor{blue!50} 1.0 &  \cellcolor{blue!30} 0.59 & \cellcolor{blue!23} 0.47 \\
			\hline
			Population Density & \cellcolor{blue!25} 0.51 & \cellcolor{blue!16} 0.32 & \cellcolor{blue!30} 0.61 & \cellcolor{blue!16} 0.32 \\
			\hline
			Num. of NP Stores & \cellcolor{blue!24} 0.48 & \cellcolor{blue!23} 0.47 & \cellcolor{blue!22} 0.45 & \cellcolor{blue!50} 1.0 \\
			\hline
			GDP per Inhabitant & \cellcolor{blue!12} 0.23 & \cellcolor{blue!05} 0.11 & \cellcolor{blue!12} 0.24 & \cellcolor{blue!1} 0.09 \\	
		\end{tabular}
	}
\end{table}

\begin{figure}[htbp]
	\centering
	\includegraphics[width=0.75\columnwidth]{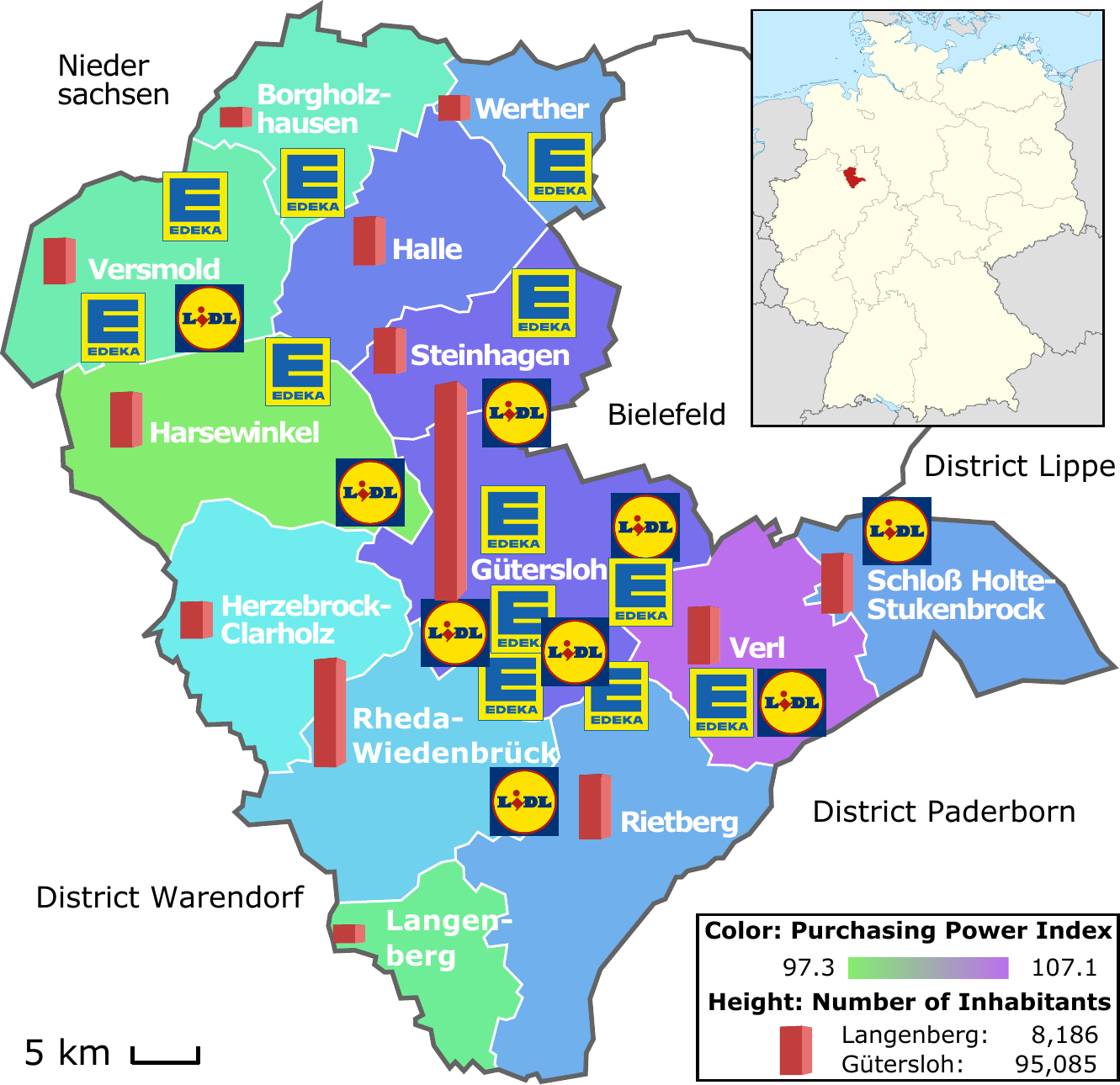}
	\caption{Distribution of Supermarket Locations within the District \emph{G\"utersloh}. Municipalities \& Cities are colored by Purchasing Power Index. Height of Bars visualize the Numb. of Inhabitants.}
	\label{fig:Municipalities_in_GT}
\end{figure}

An example case in detail is given in Figure \ref{fig:Municipalities_in_GT} where the district \emph{G\"utersloh} is visualized together with the number of inhabitants per municipality respective city and their purchasing power index. The municipality \emph{Herzebrock-Clarholz} has 15,969 inhabitants and an purchasing power index of 100. This location is among the recommended sites for a new supermarket as it fulfills the requirements. Additionally, there are no supermarkets in the close surrounding available.

\subsection{Store Distribution in Depth}
\label{subsec:distribution}

\begin{figure}[htbp]
	\centering
	\includegraphics[width=\columnwidth]{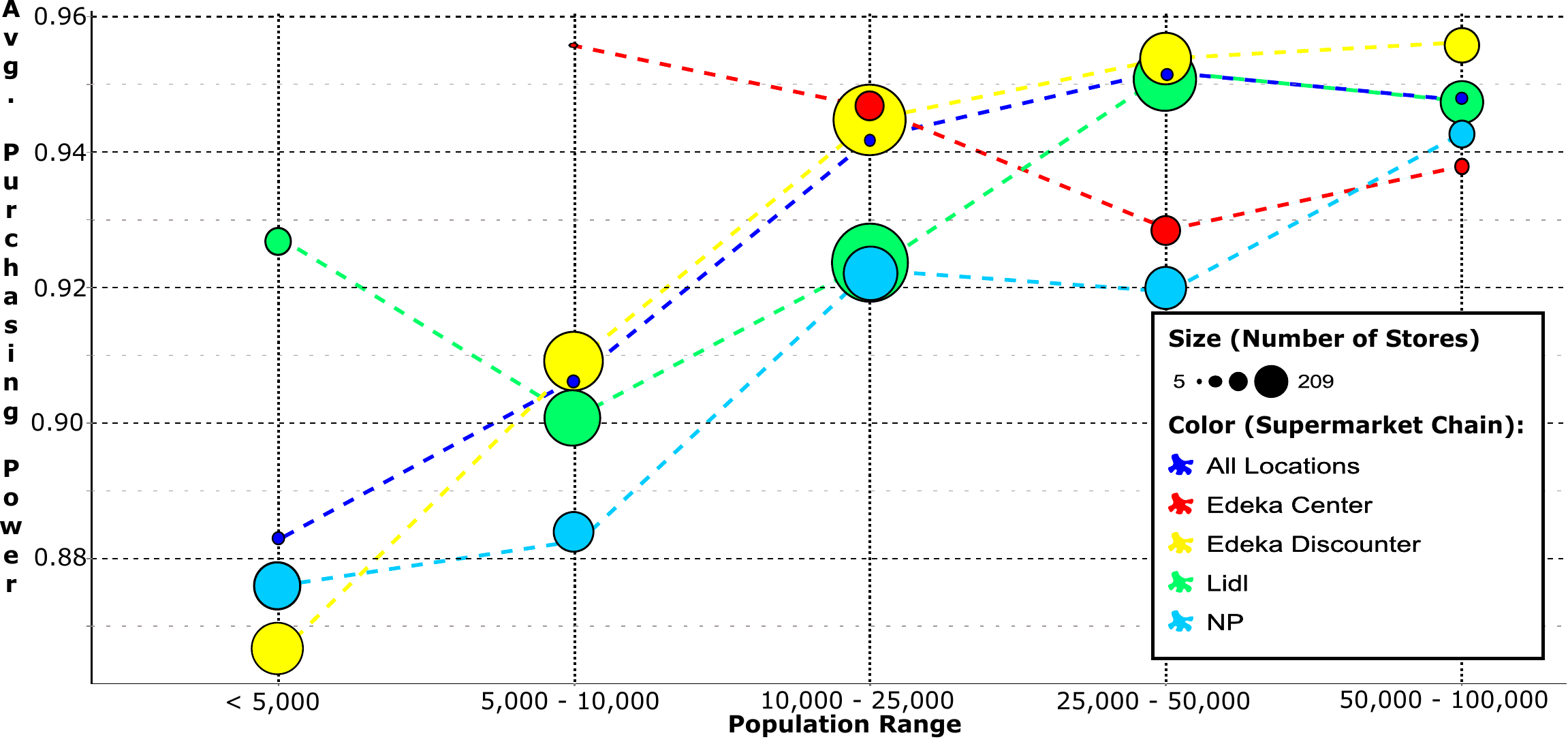}
	\caption{Distribution of Supermarket Chains according to the Average Purchasing Power Index and Grouped by Population.}
	\label{fig:StadtLand}
\end{figure}

Finally, the question is: Why have the supermarket chains chosen these sites among the places which also suit their requirements? In order to investigate the point, the decisive LF as outputted by the correlation analysis were used to examine the companies strategies in detail.
\newline
The average purchasing power\footnote{The purchasing power index describes the purchasing power of a certain region per inhabitant in comparison to the national average which itself gets the standard value of 1. 
} was investigated in Figure \ref{fig:StadtLand}. The visualization indicates that \emph{NP} focuses their stores on smaller municipalities with lower purchasing power, whereas \emph{Lidl} concentrates on bigger cities. The strategy of \emph{Edeka} is most likely to be available everywhere with medium purchasing power. The huge \emph{E-Center} stores are mainly located in cities above 10,000 inhabitants.

\begin{figure}[htbp]
	\centering
	\includegraphics[width=\columnwidth]{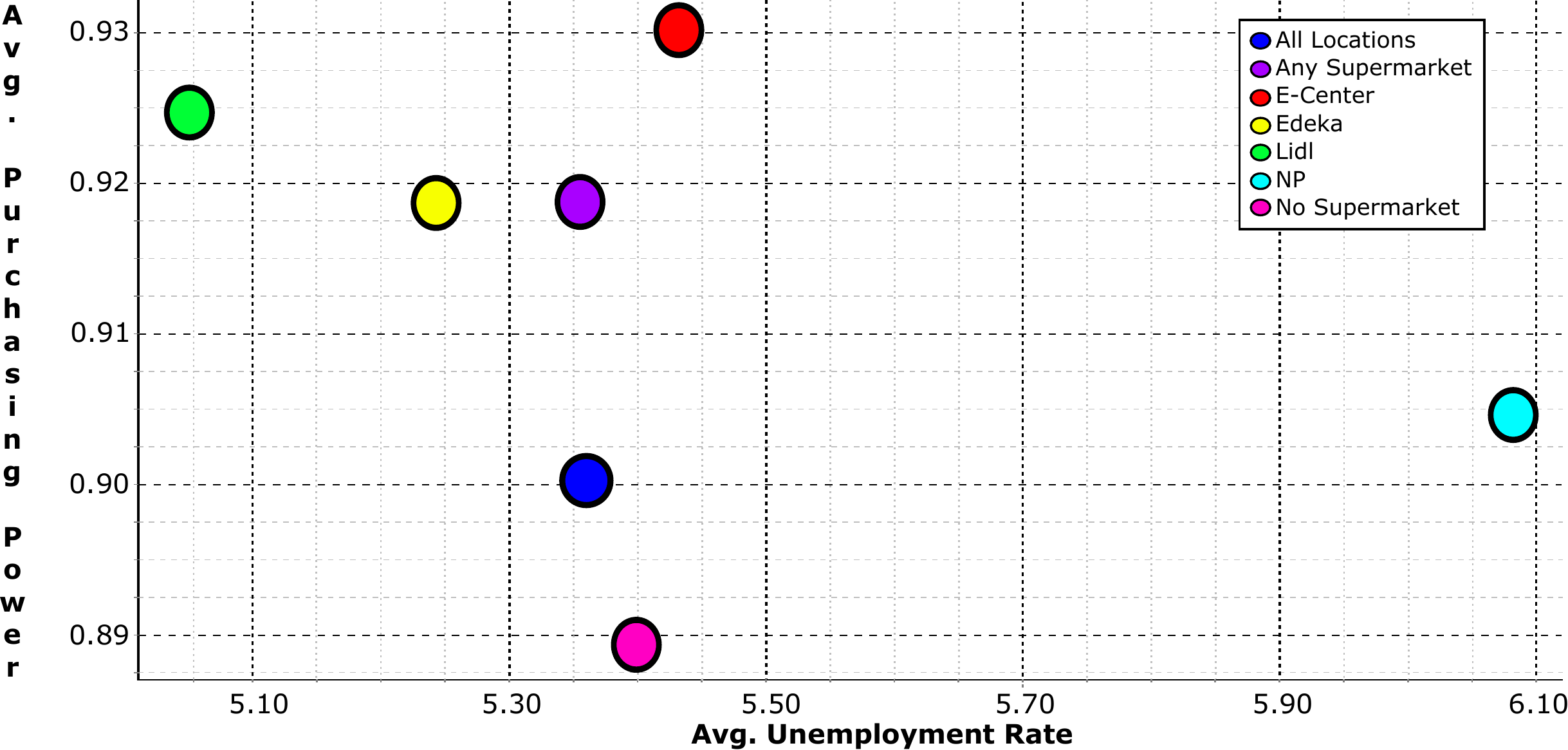}
	\caption{Distribution of Supermarket Chains according to the Average Purchasing Power Index and Unemployment Rate.}
	\label{fig:UnemploymentXPurchasePower}
\end{figure}

Figure \ref{fig:UnemploymentXPurchasePower} shows the average purchasing power index and the average unemployment rate for all municipalities where at least one supermarket of the given chain is located. Thus, the average is presented for \emph{NP} (light blue), \emph{Lidl} (green), \emph{Edeka} (yellow), and \emph{E-Center} (red). As regards the comparison of all possible sites, locations with (purple) and without (pink) supermarket of any chain as well as the all locations together (blue) are also presented.
\newline
It can be observed that \emph{Lidl}, \emph{Edeka}, and \emph{E-Center} are focusing locations with higher than average purchasing power and lower than average unemployment rate. In contrast, \emph{NP} opens stores at sites with lower than average purchasing power and higher than average unemployment rate. The Wilcoxon rank sum test (with continuity correction) confirmed that these findings are statistical significant. The population mean for the purchasing power index ranks differ for the different supermarket chains and the population ranges as presented in Figure \ref{fig:StadtLand}).
\newline
This can be explained with the two different categories of stores in food retailing: supermarkets and discount stores. Discount stores (like \emph{NP} and to certain extend \emph{Lidl}) aim to sell products at prices lower than normal supermarkets (\emph{Edeka} and \emph{E-Center}) with a focus on price rather than service, display, or wide choice. Thus, supermarkets have to chose sites with higher purchasing power as they are selling more expansive products than discount stores, which can occupy the market segment of places with lower available money as their products are less expansive.

\section{Conclusion and Future Work}
\label{sec:conclusion}

In this paper, a novel approach for data-driven site selection is proposed integrating Big Data into the decision making process of companies. It combines a data model with an enhanced recommendation, system which utilizes the existing knowledge in this context. In contrast to economic methods, the system does not need manual analysis or expert knowledge and additionally, it is capable of handling all available information about sites. More than 200 different attributes for all 11,162 municipalities in Germany have been aggregated and analyzed. The evaluation of supermarkets in Germany shows that there is a big overlap (86.4 \%) between existing stores and sites recommended by the proposed methods. In addition, the method also recommends many sites (328) for supermarket where a store should be opened. Furthermore, decisive locational factors like \textit{purchasing power} were revealed, which have an impact on the existence of supermarkets. In the future, information about the quality of existing sites will be investigated in order to learn good and poor locations as well as further studies together with economist.





\bibliographystyle{IEEEtran}
\bibliography{./literatur}
%

\end{document}